\documentclass[letterpaper, 10 pt, journal, twoside]{IEEEtran}  

\usepackage{amsmath} 
\usepackage{amssymb}  
\usepackage{xcolor}
\usepackage[hidelinks]{hyperref}

\usepackage{amsthm}
\makeatletter
\def\thm@space@setup{%
 \thm@preskip=\parskip \thm@postskip=0pt
}
\makeatother
\usepackage{thmtools}

\declaretheoremstyle[%
  spaceabove=0pt,
  spacebelow=0pt,
  headfont=\normalfont\itshape,%
  postheadspace=1em,%
  qed=\qedsymbol%
]{mystyle}
\declaretheorem[name={Proof Sketch},style=mystyle,unnumbered,
]{ps}

\declaretheoremstyle[%
  spaceabove=0pt,
  spacebelow=0pt,
  headfont=\normalfont\itshape,%
  postheadspace=1em,%
  qed=\qedsymbol%
]{mystyle}
\declaretheorem[name={Proof},style=mystyle,unnumbered,
]{myproof}

\newtheorem{thm}{Theorem}
\newtheorem{cor}[thm]{Corollary}
\theoremstyle{definition}
\newtheorem{dfn}{Definition}

\usepackage{graphicx}

\usepackage[ruled,linesnumbered,vlined]{algorithm2e}
\SetAlFnt{\fontsize{8}{9}\selectfont}
\SetAlCapFnt{\fontsize{8}{9}\selectfont}
\SetAlCapNameFnt{\fontsize{8}{9}\selectfont}
\SetArgSty{textrm}
\SetInd{0.1em}{0.5em}
\SetCommentSty{emph}
\SetAlgoSkip{}
\usepackage{etoolbox}
\makeatletter
\patchcmd{\@algocf@start}
  {-1.5em}
  {0pt}
  {}{}
\makeatother

\usepackage{cleveref}
\usepackage[skip=0pt,labelformat=simple]{subcaption}
\DeclareCaptionFont{tiny}{\footnotesize}
\usepackage{makecell}
\usepackage{booktabs}
\usepackage{multirow}
\usepackage{multicol}
\usepackage{makecell}

\begin{document}

\title{Double-Deck Multi-Agent Pickup and Delivery: Multi-Robot Rearrangement in Large-Scale Warehouses
}

\author{Baiyu Li and Hang Ma%
\thanks{Manuscript received: November 21, 2022; Revised March 2, 2023; Accepted April 18, 2023.}
\thanks{This paper was recommended for publication by Editor M. Ani Hsieh upon evaluation of the Associate Editor and Reviewers' comments.
This work was supported by NSERC under grant number RGPIN2020-06540, Huawei Tech. Canada, and a CFI JELF award. Video: \url{https://youtu.be/6CUdSmQYub8}.}
\thanks{The authors are with School of Computing Science, Simon Fraser University, Burnaby, BC V5A 1B5, Canada {\tt\footnotesize \{baiyu\_li, hangma\}@sfu.ca}}%
\thanks{Digital Object Identifier (DOI): see top of this page.}
}

\markboth{IEEE Robotics and Automation Letters. Preprint Version. Accepted April, 2023}{LI \MakeLowercase{\textit{et al.}}: DOUBLE-DECK MULTI-AGENT PICKUP AND DELIVERY: MULTI-ROBOT REARRANGEMENT IN LARGE-SCALE WAREHOUSES}

\maketitle

\begin{abstract}
We introduce a new problem formulation, Double-Deck Multi-Agent Pickup and Delivery (DD-MAPD), which models the multi-robot shelf rearrangement problem in automated warehouses. DD-MAPD extends both Multi-Agent Pickup and Delivery (MAPD) and Multi-Agent Path Finding (MAPF) by allowing agents to move beneath shelves or lift and deliver a shelf to an arbitrary location, thereby changing the warehouse layout.
We show that solving DD-MAPD is NP-hard. To tackle DD-MAPD, we propose MAPF-DECOMP, an algorithmic framework that decomposes a DD-MAPD instance into a MAPF instance for coordinating shelf trajectories and a subsequent MAPD instance with task dependencies for computing paths for agents. We also present an optimization technique to improve the performance of MAPF-DECOMP and demonstrate how to make MAPF-DECOMP complete for well-formed DD-MAPD instances, a realistic subclass of DD-MAPD instances. Our experimental results demonstrate the efficiency and effectiveness of MAPF-DECOMP, with the ability to compute high-quality solutions for large-scale instances with over one thousand shelves and hundreds of agents in just minutes of runtime.
\end{abstract}


\section{Introduction}

\IEEEPARstart{T}{he} real-world applications of multi-robot systems often require coordination between multiple agents to rearrange objects in shared environments. Warehouse robots in fulfillment centers, such as those described in \cite{kiva}, are one such example where agents are utilized to relocate inventory shelves.
In these scenarios, it is crucial for agents to avoid collisions with each other and with the objects they are relocating. Other instances of multi-agent object transportation problems include automated container relocation and 3D automated warehouse fulfillment, where the objects can be manipulated and moved by agents. In these cases, it is imperative to prevent collisions between both agents and objects.

Much research has been done on the topic of multi-robot rearrangement of inventory shelves in automated warehouses, with a focus on the simplified Multi-Agent Path Finding (MAPF) problem \cite{SternSOCS19}. In MAPF, each agent must move from its predefined start location to its predefined goal location quickly without colliding with others. Multi-Agent Pickup and Delivery (MAPD) \cite{MaAAMAS17} extends MAPF to a more realistic setting where there are more tasks than agents. In MAPD, each task has a predefined pickup location and a predefined delivery location. Each agent needs to get assigned a task and complete it by moving first to its pickup location and then to its delivery location. Task assignment and path finding are repeated until all tasks are completed. While MAPD is more practical for rearranging shelves than MAPF, it assumes a fixed storage layout for shelves and that shelves can only be picked up and delivered to designated locations. MAPD algorithms do not coordinate shelf movement explicitly, making them impossible to solve problems such as exchanging the locations of two shelves using a single agent.

\begin{figure}
    \centering
    \includegraphics[height=60pt]{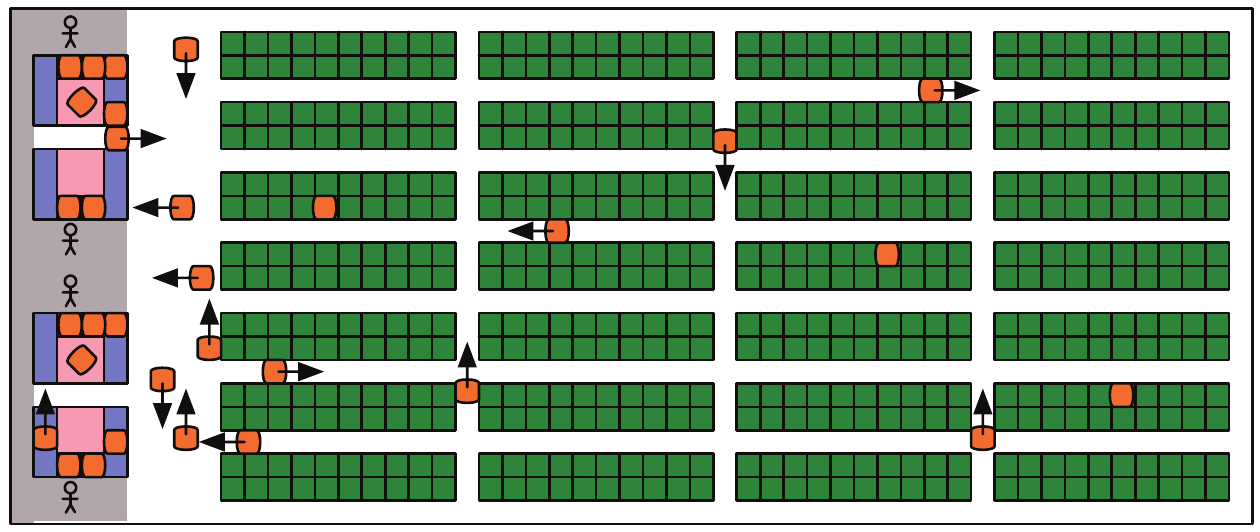}
    \caption{The typical layout of a fulfillment center \cite{kiva}.}
    \label{fig_sim}
\end{figure} 

Therefore, we introduce a novel problem formulation, Double-Deck Multi-Agent Pickup and Delivery (DD-MAPD), which extends the existing MAPF and MAPD techniques toward practical large-scale real-world warehouse autonomy. DD-MAPD addresses the multi-robot shelf rearrangement problem by allowing agents to either move beneath shelves or lift, carry, and place shelves at new locations. The name ``Double-Deck'' reflects the requirement to avoid collisions on two levels: on the high level, between shelves when they are being carried by agents, and on the low level, between agents themselves. The problem of DD-MAPD is to task $N$ agents with moving $M$ shelves from their given pickup locations to their given delivery locations, thereby changing the overall arrangement of the shelves.

From a practical perspective, the problem formulation of DD-MAPD enables the use of robots to dynamically adapt warehouse layouts in response to changing product demands. As shown in Figure 1, a typical layout of an Amazon fulfillment center features green cells representing designated storage locations for shelves, which are arranged in $4\times 7$ blocks, each consisting of $10\times 2$ storage locations. DD-MAPD allows for the creation of solutions that utilize various numbers of agents to efficiently (re)arrange shelves, such as those that are less frequently requested, into dense configurations
or selectively ``dig out'' desired shelves from densely packed blocks by moving other shelves out of the way. This approach can significantly reduce land usage, ultimately reducing the cost of an automated warehouse.

\subsection{Related Work}

\noindent\textbf{MAPF:}
MAPF is a special case of DD-MAPD where $M$ shelves are carried by $N=M$ agents.
MAPF is NP-hard to solve optimally \cite{YuLav13AAAI,surynek2010optimization} on general graphs, planar graphs \cite{Yu16RAL}, and even 2D 4-neighbor grids \cite{banfi2017intractability}. Recent MAPF solvers include reduction-based \cite{LamBHS19,Surynek19,gange2019lazy,GomezHB20}, rule-based \cite{OkumuraMDT19}, and search-based  \cite{DBLP:journals/ai/SharonSFS15,MStar,LiAAAI21a,LiAIJ21} methods.

\noindent\textbf{MAPD:}
Existing MAPD algorithms \cite{MaAAMAS17,LiuAAMAS19} decompose a MAPD instance into a sequence of task-assignment and MAPF instances. They assign tasks and plan paths for the agents whenever there is a change in the system, such as agents finishing tasks or new tasks being added. 
MAPD problems for tasks with temporal constraints \cite{GTAPF} or predefined dependencies \cite{brown2020optimal} have also been studied. However, these problems do not model shelves as movable objects that occupy locations, which is necessary for shelf rearrangements.

\noindent\textbf{Goal Sequencing and Configurable Environments:}
Recent studies \cite{surynek2021multi,ZhangAAMAS22,Ren-2022-131915} have explored MAPF variants where each agent must visit multiple goals and computes the order in which it visits the goals. DD-MAPD also requires sequencing tasks, but the task locations must be computed dynamically. \cite{bellusci2020multi} studies a MAPF variant where the warehouse layout can be changed by a blackbox, but DD-MAPD requires computing a plan for agents to change the warehouse layout.

\noindent\textbf{Plan Execution:} MAPF and MAPD plans can be executed by real robots, using a dependency graph that respects the precedence constraints \cite{honig2019persistent}. The plan execution is guaranteed to be collision-free, even with unmodeled kinematic constraints \cite{HoenigICAPS16} or delay uncertainties \cite{MaAAAI17}. Unlike our proposed framework for solving DD-MAPD, each agent has its own path in the computed plan and cannot execute other paths.

\subsection{Contributions}

We propose a novel problem formulation, DD-MAPD, to model the multi-robot shelf rearrangement problem in a warehouse. Theoretically, our work generalizes existing inapproximability and NP-hardness results of MAPD to DD-MAPD and establishes a set of sufficient conditions, namely well-formedness, for solvability. Our main contribution is a new algorithmic framework, MAPF-DECOMP, which solves a DD-MAPD instance with $N$ agents and $M$ shelves by decomposing it into an $M$-agent MAPF instance, followed by a subsequent $N$-agent MAPD instance with task dependencies---an extension to MAPD that has received limited attention thus far. MAPF-DECOMP first solves the MAPF instance to plan collision-free trajectories for the shelves. It then converts the computed shelf trajectories into tasks and solves the $N$-agent MAPD instance to assign tasks and plan paths for the agents to complete all the tasks. This decomposition has a two-fold advantage. Firstly, it enables better scalability by reducing the state space and number of agents in the planning problem of $N\times M$ agent-shelf pairs. Secondly, it leverages the advancements of off-the-shelf MAPF solvers to speed up DD-MAPD solving. We also propose an optimization technique to improve the effectiveness of MAPF-DECOMP and a variant of it that solves all well-formed DD-MAPD instances, a realistic subclass of DD-MAPD instances. Our experimental results show that MAPF-DECOMP can compute high-quality solutions for up to 1,843 shelves and 400 agents in minutes.

\section{Problem Definition}

A DD-MAPD instance consists of $N$ agents $a_1, \ldots, a_N$, $M$ shelves $s_1, \ldots, s_M$, and a connected undirected graph $G=(V, E)$, whose vertices $V$ represent locations and edges $E$ represent the connections between these locations that the agents can move along. We consider only interesting DD-MAPD instances where $M\ge N$ since, otherwise, one can use $N=M$ agents to each carry a unique shelf.

Let $\pi_i(t)$ denote the location of agent $a_i$ at timestep $t$. Each agent $a_i$ starts at its start location at timestep 0 and moves to an adjacent location or waits in its current location at each timestep. Each shelf $s_j$ starts in its pickup location $p_j$ at timestep 0 and is given a delivery location $d_j$. If a shelf $s_j$ does not need to be relocated, then $p_j=d_j$. An agent can move beneath a shelf when not carrying any shelf, and it can lift a shelf when it is in the same location as the shelf, carry the shelf from then on, and place (put down) the shelf when it arrives in another location. Agents are \emph{active} when they are carrying shelves, and \emph{free} when they are not. We assume that the time required to perform a lift or place action is 0, but our framework can easily be generalized to accommodate lift and place actions with non-zero time costs. Let $\eta_j(t)$ denote the location of shelf $s_j$ at timestep $t$. Shelves can only move when carried by agents.
There should be no collisions either between agents (on the virtual low-level deck) or between shelves (on the virtual high-level deck). A vertex collision between agents $a_i$ and $a_{i'}$ occurs iff $\pi_i(t)=\pi_{i'}(t)$; an edge collision occurs iff $\pi_i(t)=\pi_{i'}(t+1)$ and $\pi_i(t+1)=\pi_{i'}(t)$. Similarly, a vertex collision between shelves $s_j$ and $s_{j'}$ occurs iff $\eta_j(t)=\eta_{j'}(t)$; an edge collision occurs iff $\eta_j(t)=\eta_{j'}(t+1)$ and $\eta_j(t+1)=\eta_{j'}(t)$. 

The problem of DD-MAPD aims to compute collision-free paths for the agents to transport all shelves from their pickup locations to their delivery locations. The completion time of agent $a_i$ is the earliest timestep when the agent has arrived in the last location of its path and stopped moving. We use two metrics to measure the effectiveness of a DD-MAPD algorithm: the \emph{makespan}, defined as the maximum of the completion times of all agents, and the \emph{flowtime}, defined as the sum of the completion times of all agents.

\begin{figure}[t]
    \centering
    \includegraphics[width=2.67in]{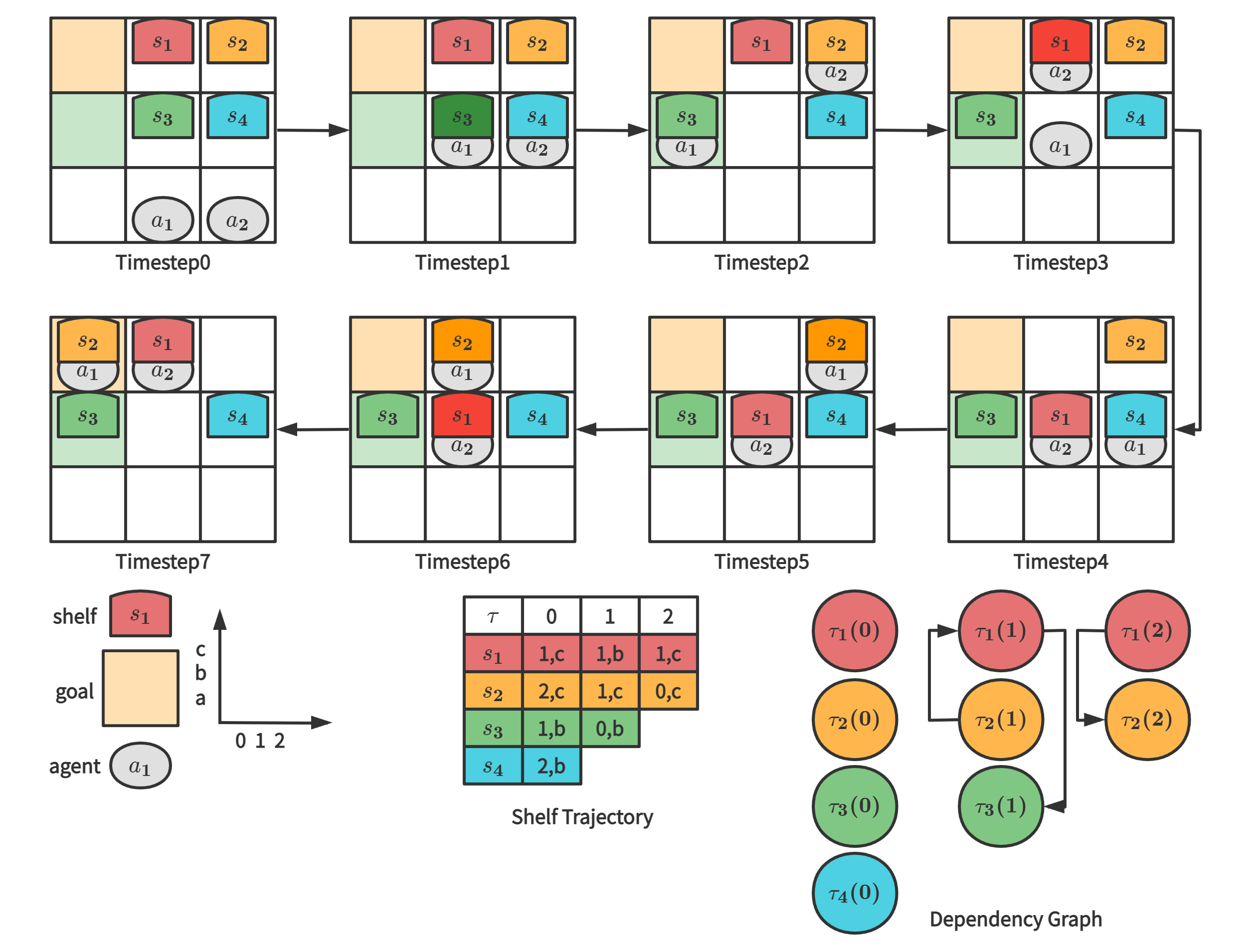}
    \caption{Example DD-MAPD instance with two agents and four shelves on a 2D 4-neighbor grid. Shelves $s_2$ and $s_3$ need to be relocated to the orange and green cells, respectively. Other shelves do not need to be relocated. Top: Locations of agents and shelves at each timestep. Bottom: Shelf trajectories and the resulting dependency graph.}
    \label{fig_exp}
\end{figure}

We use the DD-MAPD instance shown in \Cref{fig_exp} as our running example. \Cref{fig_exp} (Top) demonstrates a solution with makespan 7 and flowtime 14 ($=7+7$).

\section{Complexity and Solvability}
We now generalize existing complexity results for MAPD to DD-MAPD and identify sufficient conditions that make DD-MAPD instances solvable.

\subsection{Complexity}

We first show a constant-factor inapproximability result for DD-MAPD with respect to makespan minimization by reusing the reduction \cite{MaThesis20} from the NP-complete $\le$3,$=$3-SAT problem \cite{cat1984} to MAPD, similar to that for MAPF \cite{MaAAAI16}.

\begin{thm}\label{thm:makespan_hardness}
    For any $\epsilon > 0$, it is NP-hard to find a $(4/3 - \epsilon)$-approximate solution to DD-MAPD for makespan minimization.
\end{thm}

\begin{ps}
    We use the same reduction as that used in the proof of Theorem 4.5 in \cite{MaThesis20} to construct a DD-MAPD instance with $M=N=2\mathcal{N}+\mathcal{M}$ shelves and agents for a given $\le$3,$=$3-SAT instance with $\mathcal N$ variables and $\mathcal M$ clauses. The arguments for MAPD still hold for DD-MAPD since each constructed agent carries only its corresponding constructed shelf. The constructed DD-MAPD instance has a solution with makespan three iff the $\le$3,$=$3-SAT instance is satisfiable, and always has a solution with makespan four, even if the $\le$3,$=$3-SAT instance is unsatisfiable.
\end{ps}

The constructed DD-MAPD instance in the above proof has the property that the completion time of each agent is at least three. Therefore, if the makespan is three, then every agent completes in exactly three timesteps, and the flowtime is $3N$ ($= 3(2\mathcal{N}+\mathcal{M})$). Moreover, if the makespan exceeds three, then the flowtime exceeds
$3M$, yielding the following corollary:

\begin{cor}
It is NP-hard to find an optimal solution to DD-MAPD for flowtime minimization.
\end{cor}

\subsection{Baseline Complete Algorithm for Well-Formed Instances}
We characterize a subclass of solvable DD-MAPD instances, called \emph{well-formed} DD-MAPD instance, that generalize well-formed MAPD instances \cite{MaAAMAS17}, even though we often need to solve non-well-formed DD-MAPD instances in practice. We first define that a MAPF solution for shelves (by treating shelves as agents that can move by themselves) is \emph{1-robust} \cite{atzmon2020robust} iff, at any time step, a shelf is not allowed to (move to and) occupy a location at the next time step if the location is currently occupied. We note that the requirement for the existence of a 1-robust MAPF solution is not overly restrictive in practice since a sufficient condition for it is that at least two vertices of the MAPF graph are unoccupied \cite{PushAndSwap}.%
\begin{dfn}[Well-formedness and safe 1-robustness]
A DD-MAPD instance is \emph{well-formed} iff all agents start at different locations, graph $G$ remains connected if the start locations of any $N-1$ agents are removed, and there exists a \emph{safe} 1-robust MAPF solution for shelves, defined as one that does not use the start location of any agent.
\end{dfn}

We then sketch a baseline algorithm that is complete for all well-formed DD-MAPD instances as follows: It first generates trajectories for all shelves by computing a safe 1-robust MAPF solution for them; It then uses any single agent to execute all shelf trajectories in locked steps, namely proceeding to the next step only after executing one step of the MAPF solution for all shelves, while letting all other agents wait at their start locations. It is a straightforward observation that all well-formed DD-MAPD instances are solvable, and this baseline algorithm solves all of them. Our framework, MAPF-DECOMP, is similar to this baseline algorithm in that it first computes the shelf trajectories, but it differs in that it utilizes multiple agents to execute them.

\section{MAPF-DECOMP}

\Cref{MAPF-DECOMP} shows the pseudocode of MAPF-DECOMP.
The algorithm starts by calling a MAPF solver to compute collision-free trajectories $\tau$ for all shelves from their pickup locations to delivery locations [\Cref{a11}]. These trajectories represent the intended path for each shelf, but the actual path of each shelf is executed by the agents. To distinguish the trajectories from the actual paths of the shelves or agents, we refer to $\tau$ as \emph{trajectories}.\footnote{This paper adopts a non-standard use of the terms ``paths'' and ``trajectories'' by reversing their conventional definitions. The purpose of this is to maintain consistency with the usage of ``paths'' in the MAPF literature, whereas in robotics, a ``trajectory'' typically refers to the path/executed followed by an agent as a function of time.}
Next, MAPF-DECOMP converts the trajectories into a dependency graph that implicitly partitions each trajectory into segments [\Cref{a12}]. Each segment represents a portion of the trajectory that can be executed by an agent by carrying the shelf and following the segment. Finally, MAPF-DECOMP solves a specialized MAPD instance with task dependencies by assigning shelves (segments) to agents and planning paths for them to complete all executable segments at each timestep where an agent changes from active to free or vice versa, while respecting the dependencies between segments [Lines \ref{a12.5}-\ref{a15}]. Unlike existing MAPD algorithms, MAPF-DECOMP does not plan paths for active agents but lets them follow the planned trajectories of the shelves they carry. The paths of active agents are thus the unexecuted portion of the trajectories. 

\begin{algorithm}[t]
    \caption{MAPF-DECOMP\label{MAPF-DECOMP}}
    \linespread{0.8}\selectfont
    $\tau\leftarrow$ trajectories of shelves by MAPF\;\label{a11}
    $\mathcal{G}=(\mathcal{V},\mathcal{E})\leftarrow \textnormal{BuildDep}()$\;\label{a12}
    \ForEach{agent $\mathit{a_i}$\label{a12.5}}{
        $\mathit{States_i.type \gets free}$,
        $\mathit{States_i.shelf \gets null}$\;\label{initial shelf}
    }
    \ForEach{shelf $s_j$}{
        $\mathit{Steps_j\leftarrow 0}$\;\label{initial step}
    }
    \While{there are uncompleted shelves}{
        $\textnormal{Update}\mathit{(States,\mathcal{G})}$\;\label{a13}
        \If{ $\mathit{States}$ has changed}{
            $\mathit{FreePaths \gets\textnormal{AssignAndPlan}(States)}$\;\label{a14}
        }
        System proceeds to the next timestep\;\label{a15}
        $\mathit{\textnormal{Move}(States,\tau, FreePaths)}$\;\label{a16}
    }
\end{algorithm}

To solve the specialized MAPD instance resulting from the dependency graph, MAPF-DECOMP maintains the current state of each agent $a_i$ in the variable $\mathit{States}_i$ that consists of two attributes: $\mathit{States}_i.\mathit{type}$ is either $\mathit{free}$ or $\mathit{active}$ and initially set to $\mathit{free}$; $\mathit{States}_i.\mathit{shelf}$ is the shelf assigned to agent $a_i$ and initially set to $\mathit{null}$ [\Cref{initial shelf}].
In addition, each shelf $s_j$ is assigned a step $s_j.\mathit{step} = k$, which represents its current location $\tau_j(k)$ according to its trajectory $\tau_j$. The step $s_j.\mathit{step}$ is initially set to 0, corresponding to the pickup location $\tau_j(0)=p_j$ [\Cref{initial step}].
We recall that a shelf is completed if it has arrived at its delivery location and remains there. At each timestep where there are uncompleted shelves, MAPF-DECOMP updates the states of all agents based on the dependency graph and the steps of all shelves [\Cref{a13}]. If any agent changes its state, MAPF-DECOMP (re)assigns shelves or safe locations to free agents and calls a MAPF solver to compute collision-free paths for them to reach their assigned destinations [\Cref{a14}].
The algorithm then proceeds to the next timestep and moves all agents and the shelves they carry to their next locations [Lines \ref{a15}-\ref{a16}].



\subsection{Building Dependency Graph}\label{sec:BuildDep}





Given the trajectories $\tau$ as the result of calling a MAPF solver, MAPF-DECOMP calls Function BuildDep() to construct a dependency graph. The trajectories $\tau$ define a total order on their entries $\tau_j(k)$, which the dependency graph relaxes to a partial order. The resulting dependency graph $\mathcal{G}=(\mathcal{V},\mathcal{E})$ is a directed graph whose vertices $\mathcal{V}$ are the trajectories entries $\tau_j(k)$.
For all $j,j',k,k'$ with $j\neq j'$, $k>k'$, and location $\tau_j(k) = \tau_{j'}(k')$, an edge $\langle \tau_j(k), \tau_{j'}(k'+1) \rangle$ exists, which represents that shelf $s_j$ should be in step $k$, namely the location specified by $\tau_j(k)$ (that follows $\tau_j(k)$, which specifies the same location as  $\tau_{j'}(k')$), no earlier than shelf $s_{j'}$ is in step $k'+1$, namely the location specified $\tau_{j'}(k'+1)$. That is, whether an agent is allowed to execute the segment starting from step $k$ or not depends on whether shelf $s_{j'}$ has been in step $k'+1$. By construction, cycles can only exist for entries $\tau_j(k)$ with the same step $k$. 
The above construction of the dependency graph is similar to the one used in the previous work \cite{HoenigICAPS16,MaAAAI17} and guarantees that the execution of the collision-free trajectories of the shelves is also collision-free. It differs from the one used in the previous work in not constructing edges between entries of the same trajectory. \Cref{fig_exp} (bottom) shows the dependency graph for our running example.


\subsection{Updating Agent States}\label{sec:Update}




    

\begin{algorithm}[t]
    \caption{$\textnormal{Update}\mathit{(States,\mathcal{G})}$\label{alg:Update}}
    \linespread{0.8}\selectfont
    {\color{black}$\mathit{\mathcal A_{sc}\leftarrow \emptyset}$}\;\label{alg2:emptyset}
    \ForEach{agent $\mathit{a_i}$}{
        $s_j\gets\mathit{States_i.shelf}$\;
        $\mathit{numDeps}\gets$ $|$outgoing edges of $\tau_j(s_j.\mathit{step}+1)$ in $\mathcal{G}|$ \label{alg2:num_deps}\;
        {\color{black}
        $\mathit{isSoftDep\leftarrow numDeps}=1$ \textbf{?} $\textnormal{SoftDep}\mathit{(States_i.shelf,\mathcal{G})}$ \textbf{:} $\mathit{false}$\;}
        \If{$\mathit{States_i.type=free}$ \textbf{and} agent $a_i$ at the same location as $\mathit{States_i.shelf}$}{\label{a21}
                {\color{black}\If{ $\mathit{isSoftDep}$}{\label{a25}
                    $\mathit{\mathcal A_{sc}\leftarrow \mathcal A_{sc} \cup a_i}$\;
                }}\label{a26}
                \ElseIf{$\mathit{numDeps}\geq 1$}{\label{a23}
                    $\mathit{States_i.shelf\leftarrow null}$\;
                }\label{a24}
                \Else{\label{a27}
                    $\mathit{States_i.type\leftarrow active}$\;
                }\label{a28}
        }
        \ElseIf{ $\mathit{States_i.type=active}$}{\label{alg2:active}
            \If{ $\mathit{States_i.shelf}$ is completed}{
                $\mathit{States_i.shelf\leftarrow null}$\;
                $\mathit{States_i.type\leftarrow free}$\;\label{alg2:become_free}
            }
            {\color{black}\ElseIf{$\mathit{isSoftDep}$\label{alg2:active_soft}}{
                $\mathit{\mathcal A_{sc}\leftarrow \mathcal A_{sc} \cup a_i}$\;\label{alg2:add_active}
            }}
            \ElseIf{$\mathit{numDeps}\ge 1$\label{alg2:active_hard}}{
                $\mathit{States_i.shelf\leftarrow null}$\;
                $\mathit{States_i.type\leftarrow free}$\;\label{alg2:become_free2}
            }
        }\label{a210}
    }
    {\color{black}
    $\mathit{\mathcal A_\mathit{noMove}\leftarrow \textnormal{FindNoMove}(\mathcal A_{sc})}$\;\label{alg2:no_move}      
    \ForEach{agent $\mathit{a_i\in \mathcal A_{sc}}$}{
        \If{ $\mathit{a_i\in \mathcal A_\mathit{noMove}}$}{
            $\mathit{States_i.shelf\leftarrow null}$\;\label{alg2:no_move_unassigned}  
            $\mathit{States_i.type\leftarrow free}$\;\label{alg2:no_move_free}  
        }
        \Else{
            $\mathit{States_i.type\leftarrow active}$\;\label{alg2:res_active}  
        }
    }}
\end{algorithm}

At each timestep, MAPF-DECOMP calls Function Update() to update the states of agents.
Update() serves three purposes: (1) It checks whether a free agent located at its assigned shelf's location can execute the shelf's trajectory. (2) It verifies if an active agent can continue executing the trajectory of its assigned shelf or if the shelf's trajectory is completed. (3) If the shelf trajectories are not 1-robust, the function checks whether the next trajectory entries of the assigned shelves of multiple agents form a cycle or path and whether the shelves can be executed simultaneously.

We define a (simple) \emph{path} on $\mathcal{G}$ as a sequence of vertices that has an outgoing edge from each vertex in the sequence to its successor in the sequence, with no repeated vertices and edges. We define a (simple) \emph{cycle} on $\mathcal{G}$ as a (simple) path except that the first and last vertices are the same.

\Cref{alg:Update} shows the pseudocode of Update(). $\mathit{\mathcal A_{sc}}$ is the set of possible agents whose assigned shelves have dependencies that might be released simultaneously in the next step. $\mathit{\mathcal A_{sc}}$ is set to empty initially [\Cref{alg2:emptyset}]. For each agent $a_i$, $\mathit{numDeps}$ stores the number of dependencies of the next step of its assigned shelf $s_j$ [\Cref{alg2:num_deps}]. If $\mathit{numDeps}=1$, let the only dependency be $\langle \tau_j(s_j.\mathit{step}+1), \tau_{j'}(k'+1)\rangle$ (with $\mathit{\tau_j(s_j.\mathit{step}+1)=\tau_{j'}(k')}$ and $\mathit{step}+1>k'$). In this case, Function SoftDep() returns $\mathit{true}$ iff $\mathit{Steps_{j'}=k'}$ (namely, shelf $s_{j'}$ is in step $k'$ that specifies the same location as the next step of shelf $s_j$) and stores the result in the Boolean variable $\mathit{isSoftDep}$. In this case, we say that shelf $s_j$ is \emph{softly constrained}, or has a \emph{soft dependency}, because both shelves $s_j$ and $s_{j'}$ can move one step forward simultaneously, which includes cases where $\tau_j(s_j.\mathit{step}+1)$ and $\tau_{j'}(k'+1)$ are in a cycle or path.
For each free agent $a_i$ that is in the same location as its assigned shelf [\Cref{a21}], 
if the shelf has a soft dependency, then agent $a_i$ is added to $\mathit{\mathcal A_{sc}}$ [Lines \ref{a25}-\ref{a26}]. Otherwise, if the shelf has dependencies (that are thus \emph{hard}), then it is unassigned from agent $a_i$ [Lines \ref{a23}-\ref{a24}]. Otherwise, the shelf has no dependency, and agent $a_i$ changes from free to active and starts executing the trajectory of the shelf [Lines \ref{a27}-\ref{a28}]. All other free agents do not change their states (remain free).
For each active agent $a_i$, if its assigned shelf is completed, then the shelf is unassigned from it, and it changes from active to free [Lines \ref{alg2:active}-\ref{alg2:become_free}]. Otherwise, if the shelf has a soft dependency, then agent $a_i$ is added to $\mathit{\mathcal A_{sc}}$ [Lines \ref{alg2:active_soft}-\ref{alg2:add_active}]. Otherwise, if the shelf has dependencies (that are thus hard), then it is unassigned from agent $a_i$, and the agent changes from active to free [Lines \ref{alg2:active_hard}-\ref{alg2:become_free2}]. All other active agents do not change their states and continue executing the trajectories of the shelves they carry.

Function Update() then calls the procedure FindNoMove() to identify the set $\mathit{\mathcal A_\mathit{noMove}}$ of any agents in $\mathit{\mathcal A_{sc}}$ that cannot move to the locations specified by the next step of the trajectories of their assigned shelves. We recall that each such shelf $s_j$ is softly constrained, namely, its next trajectory entry depends on the next trajectory entry of another shelf $s_{j'}$ (not necessarily assigned to an agent in $\mathit{\mathcal A_{sc}}$) and it can thus move one step only no earlier than shelf $s_{j'}$ has moved one step. FindNoMove() considers the subgraph $\mathcal{G}_\mathit{sc}$ of $\mathcal{G}$ induced by the next trajectory entries of shelves assigned to all agents in $\mathit{\mathcal A_{sc}}$. Each such trajectory entry is in either a cycle or a path on $\mathcal{G}_\mathit{sc}$ since it has one outgoing edge in $\mathcal{G}_\mathit{sc}$ except for the case that it is the last vertex on some path and its (only) outgoing edge in $\mathcal{G}$ points to a trajectory entry of some shelf $s_{j'}$ that is not assigned to any agent in $\mathit{\mathcal A_{sc}}$ (the edge thus does not belong to $\mathcal{G}_\mathit{sc}$). In this case, if shelf $s_{j'}$ is assigned to either a free agent (not in $\mathit{\mathcal A_{sc}}$) or no agent at all, then the outgoing edge (dependency) is not released since shelf $s_{j'}$ is not carried by any agent, The shelves with their next trajectory entries on this path cannot move, and the agents that they are assigned to are thus added to $\mathit{\mathcal A_\mathit{noMove}}$. 

\subsection{Shelf Assignment and Path Planning}\label{sec:AssignAndPlan}

MAPF-DECOMP calls Function AssignAndPlan() if Update() changes an agent's type from free to active or from active to free and unassigns its assigned shelf. AssignAndPlan() then assigns \emph{executable} shelves, namely ones that are not constrained at their current steps, to free agents and plan their paths, unless the shelves are already carried by active agents. The function operates in multiple rounds, with the aim of making more shelves executable in each subsequent round, as constraints are lifted by the paths planned in the previous round. To do so, AssignAndPlan() procedure follows these steps in each round: (1) It constructs a candidate set of unassigned and executable shelves. (2) If no such shelves exist, the function simulates Function Move() that lets all agents (including the free agents that got assigned shelves in the previous rounds and all active agents) follow their paths. During simulation, each shelf moves together with its assigned active agent until it reaches a hard dependency in its next trajectory entry. AssignAndPlan() then identifies any newly executable unassigned shelves and adds them to the candidate set. (3) If no such shelves become executable from the simulation, the function identifies all unassigned shelves whose next trajectory entries form a cycle in $\mathcal{G}$ and adds them to the candidate set since they can be executed simultaneously. Once the candidate set consists of one or more shelves, AssignAndPlan() calls the Hungarian algorithm \cite{Kuhn55thehungarian} to find a minimal-cost assignment of shelves to free agents and plans their paths. The cost of assigning a shelf to an agent is calculated as the maximum of the shortest-path distance between the agent and the shelf and the timestep when the shelf first becomes executable. AssignAndPlan() then calls a MAPF solver to compute paths for these agents from their current locations to the current locations of their assigned shelves, avoiding collision with paths of active agents and any paths of free agents already planned in the previous rounds. When all executable shelves have been assigned, for each free agent that remains unassigned, the function assigns it a unique location closest to it and not on any active agent's path. The function then calls a MAPF solver to compute paths for these unassigned agents, avoiding collisions with the paths of all other agents.
 
\subsection{Moving Agents and Shelves}\label{sec:Move}

\begin{algorithm}[t]
    \linespread{0.8}\selectfont
    \caption{$\mathit{\textnormal{Move}(States,\tau, FreePaths)}$\label{alg:Move}}
    \ForEach{agent $\mathit{a_i}$}{
        \If{ $\mathit{States_i.type=active}$}{
            $s_j \gets \mathit{States_i.shelf}$\;
            Update the location of $a_i$ and $s_j$ according to $\tau_j$\label{alg3:update_active}\;
            $\mathit{s_j.step\leftarrow s_j.step+1}$\label{alg3:update_step}\;
            Remove incoming edges of $\mathit{\tau_{j}(s_j.step)}$ from $\mathcal{G}$\label{alg3:remove_deps}\;
        }
        \Else{
            Update the location of $a_i$ according to $\mathit{FreePaths_i}$\;\label{alg3:update_step_free}
        }
    }
\end{algorithm}

When the system proceeds to the next timestep, MAPF-DECOMP calls Function Move() to move agents and shelves one step forward. \Cref{alg:Move} shows the pseudocode of Move(). Each active agent $a_i$ and the shelf it carries move one step according to the trajectory $\tau_j$ of the shelf [Lines \ref{alg3:update_active}-\ref{alg3:update_step}], which, as a result, releases the dependencies of the corresponding entry of $\tau_j$ (by removing all incoming edges of the entry from $\mathcal{G}$) [\Cref{alg3:remove_deps}]. Each free agent moves one step according to $\mathit{FreePaths_i}$ [\Cref{alg3:update_step_free}].

\subsection{Running Example}\label{sec:Example}

\Cref{fig_exp} (Top) shows the execution of the shelf trajectories for our running example. For ease of presentation, we point out only the most insightful details. At timestep 0, $a_1$ and $a_2$ are assigned $s_3$ and $s_1$ since $s_2$ is not executable in Round 1.
At timestep 1, $a_1$ changes from free to active.
At timestep 2, $a_1$ completes $s_3$ and becomes free. In Round 1, only $s_1$ is executable and is assigned to $a_2$. In Round 2, $s_2$ is assigned to $a_1$. At timesteps 4 and 5, Update() sets $a_2$ to free even though it is in the same location as its assigned shelf $s_1$ since $\langle\tau_1(2),\tau_2(2)\rangle$ is a hard constraint. At timestep 6, Update() adds only $a_2$ to $\mathit{\mathcal A_{sc}}$ and changes it to active since $\langle\tau_1(2),\tau_2(2)\rangle$ is a soft constraint. 


\subsection{Optimization: Involving Future (IVF)}\label{sec:Optimizations}

We now present an optimization technique called Involving Future (IVF) that aims to enhance the effectiveness of the shelf assignment procedure in MAPF-DECOMP. IVF improves AssignAndPlan() by considering not only the currently free agents but also those that will become free in $K$ timesteps for shelf assignment. This allows for a larger pool of agents to be considered for shelf assignment, potentially resulting in a better overall assignment. To achieve this, IVF simulates both Update() and Move() for $K$ iterations and includes active agents that change to free during the simulation in the shelf assignment and path planning process. The number of timesteps elapsed when each such agent becomes free is recorded and added to the shortest-path distance between the agent and any shelf when calculating the cost of assigning the shelf to the agent.

\section{Prioritized Planning (PP) for Completeness}\label{sec:completeness}

MAPF-DECOMP is not guaranteed to solve all well-formed DD-MAPD instances due to two reasons: (1) If the shelf trajectories are not 1-robust, multiple shelves with soft dependencies may form a cycle (as detailed in Section \ref{sec:AssignAndPlan}), which cannot be resolved if the total number of agents is smaller than the number of shelves in the cycle. (2) Function AssignAndPlan() plans paths for free agents to locations different from the start locations of agents, which does not guarantee collision-free paths from those locations exist.

Thus, we propose a variant of MAPF-DECOMP, MAPF-DECOMP(PP), that is complete for all well-formed DD-MAPD instances. This variant differs from the original in the following ways: (1) It uses a MAPF solver to compute safe 1-robust shelf trajectories. (2) In each assignment round, AssignAndPlan() assigns only one shelf from the candidate set to an agent by selecting the pair with the smallest assignment cost, and then uses a multi-label A* search \cite{grenouilleau2019multi} to compute a time-minimal path for the agent, avoiding any collisions with the paths of other agents. The path first moves the agent from its current location to the current location of the shelf, then follows the trajectory of the shelf without waiting until the shelf is constrained, and finally moves the agent to its start location. Any free agent that remains unassigned when all executable shelves have been assigned keeps its current path that ends in its start location. Therefore, each agent maintains the invariant that its path always ends in its start location, inspired by the ``reserving dummy paths'' deadlock-avoidance technique for MAPD \cite{LiuAAMAS19}.

\begin{thm}
MAPF-DECOMP(PP) solves all well-formed DD-MAPD instances. 
\end{thm}

\begin{myproof}
Since the dependency graph for 1-robust trajectories is acyclic, Function AssignAndPlan() adds at least one unassigned shelf that is or (in simulation) will become executable to the candidate set and assigns it to a free agent. The multi-label A* search guarantees to find a collision-free path for the agent to execute the shelf trajectory segment since such a path always exists. For example, the agent can first follow its old path and wait at its start location until all other agents have completed their paths and stay at their start locations indefinitely. Without passing through the start locations of other agents, the agent can then move to the current location of its assigned shelf, follow the trajectory of the shelf without waiting or causing shelf collisions until the shelf is constrained, and finally return to its start location. Thus, all shelf trajectories will eventually be executed since at least one additional trajectory segment is assigned and executed each time AssignAndPlan() is called.
\end{myproof}

\begin{figure}[t]
\centering  
\includegraphics[height=80pt]{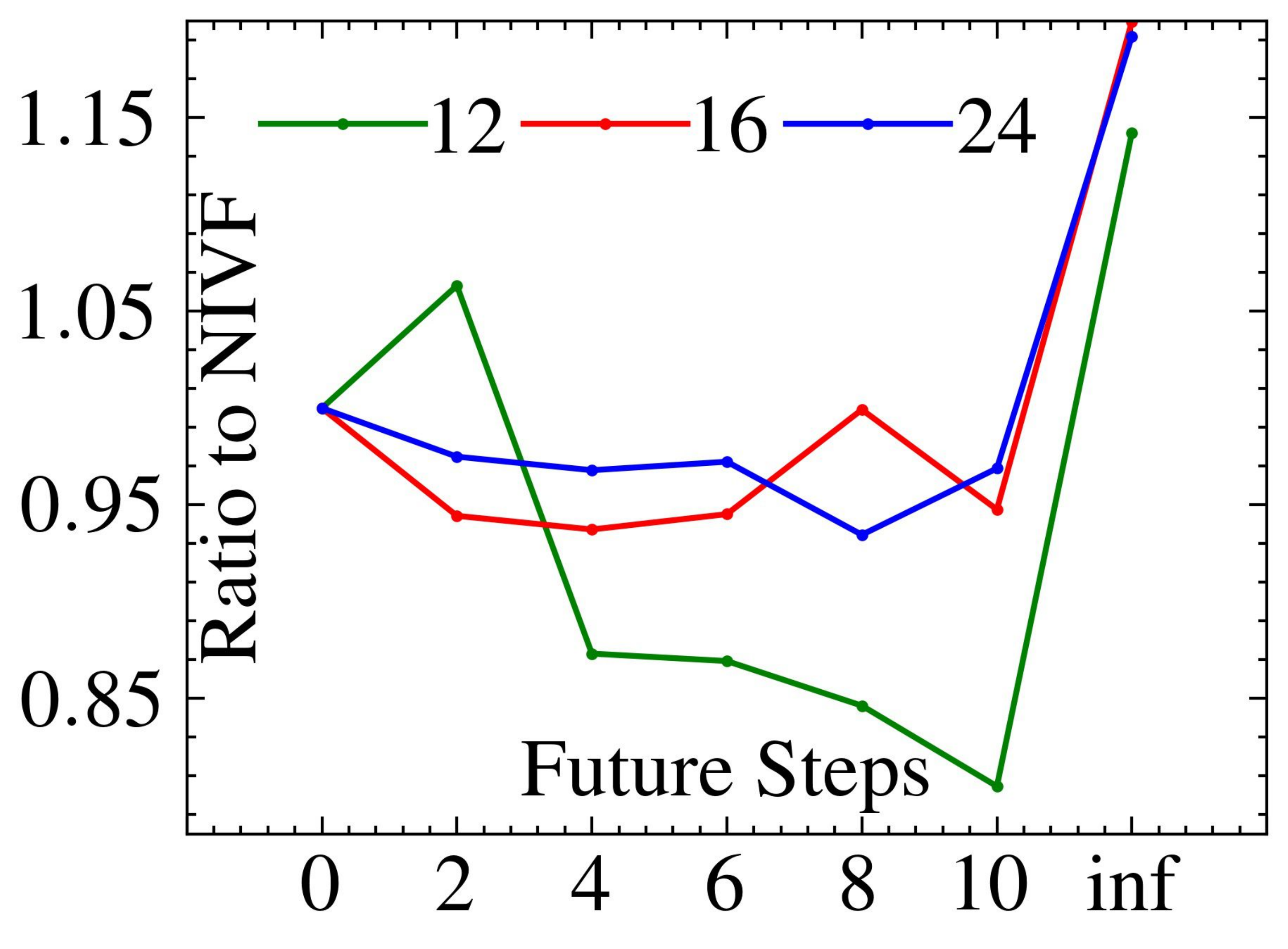}\includegraphics[height=80pt]{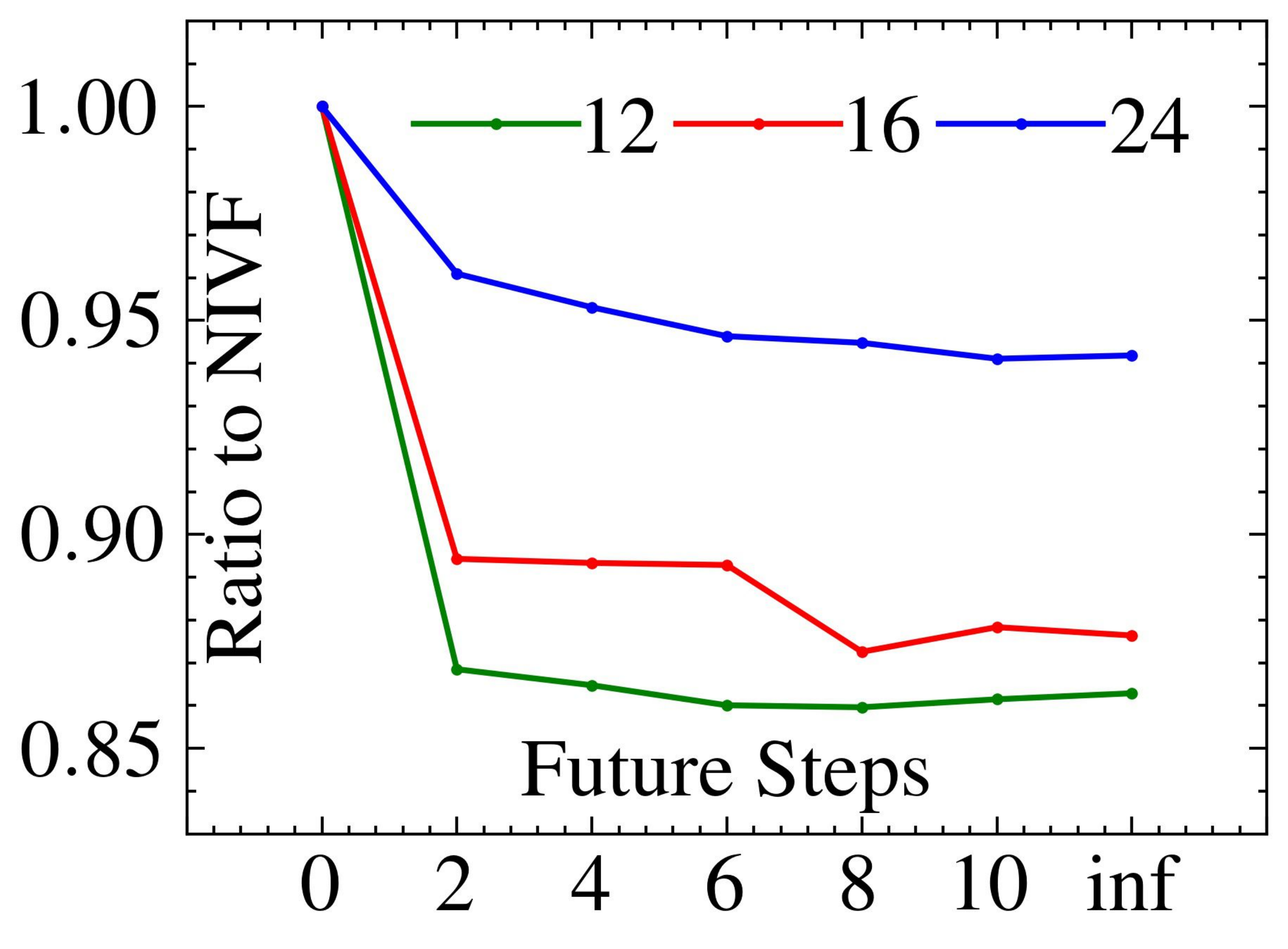}
\caption{Agent time (left) and makespan (right) ratios of IVF against NIVF for different $K$. Legends show grid sizes.}\label{fig_inv_step}
\end{figure}

\begin{figure*}[t]
    \centering
    \includegraphics[height=85pt]{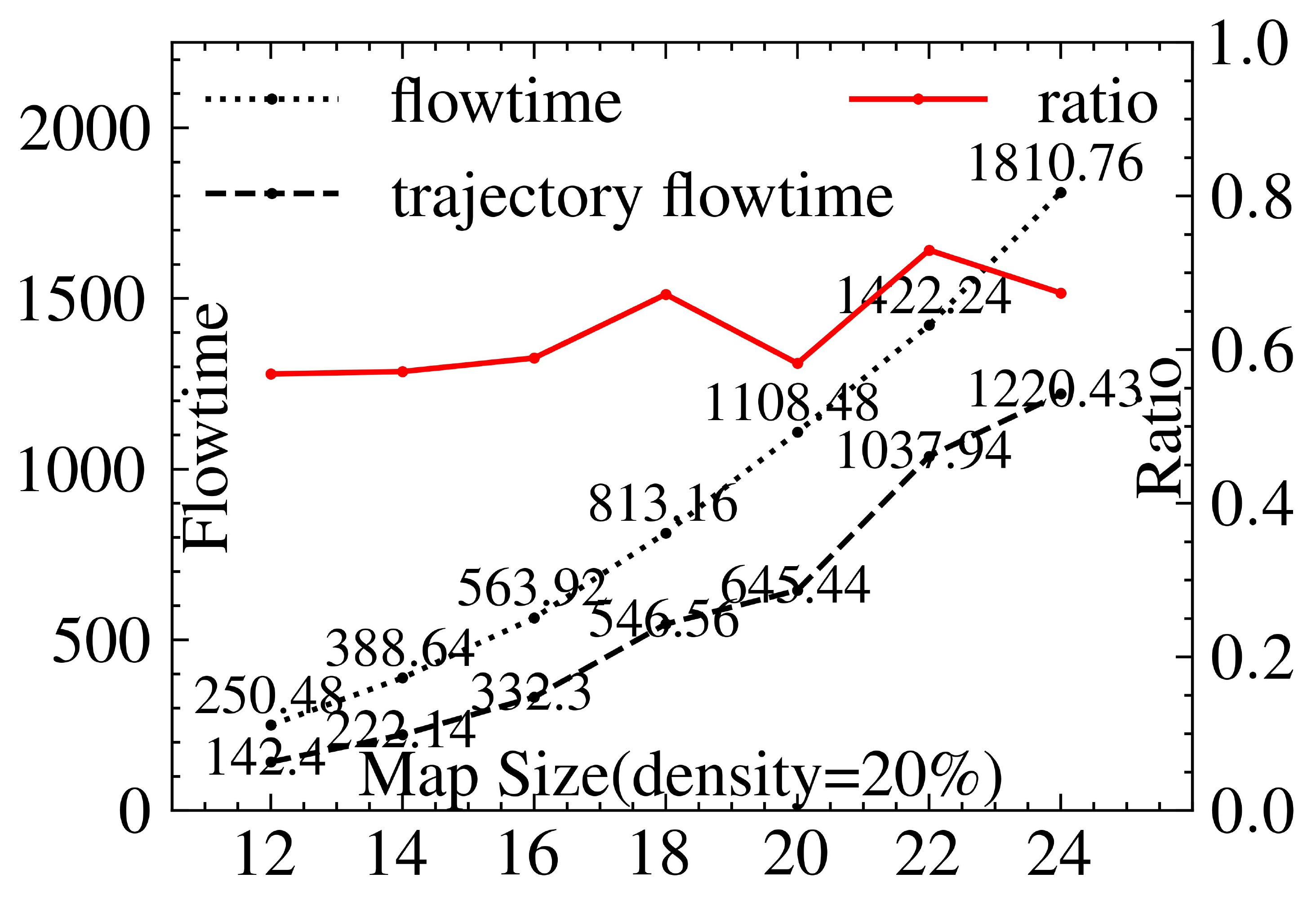}~~~~~~~~\includegraphics[height=85pt]{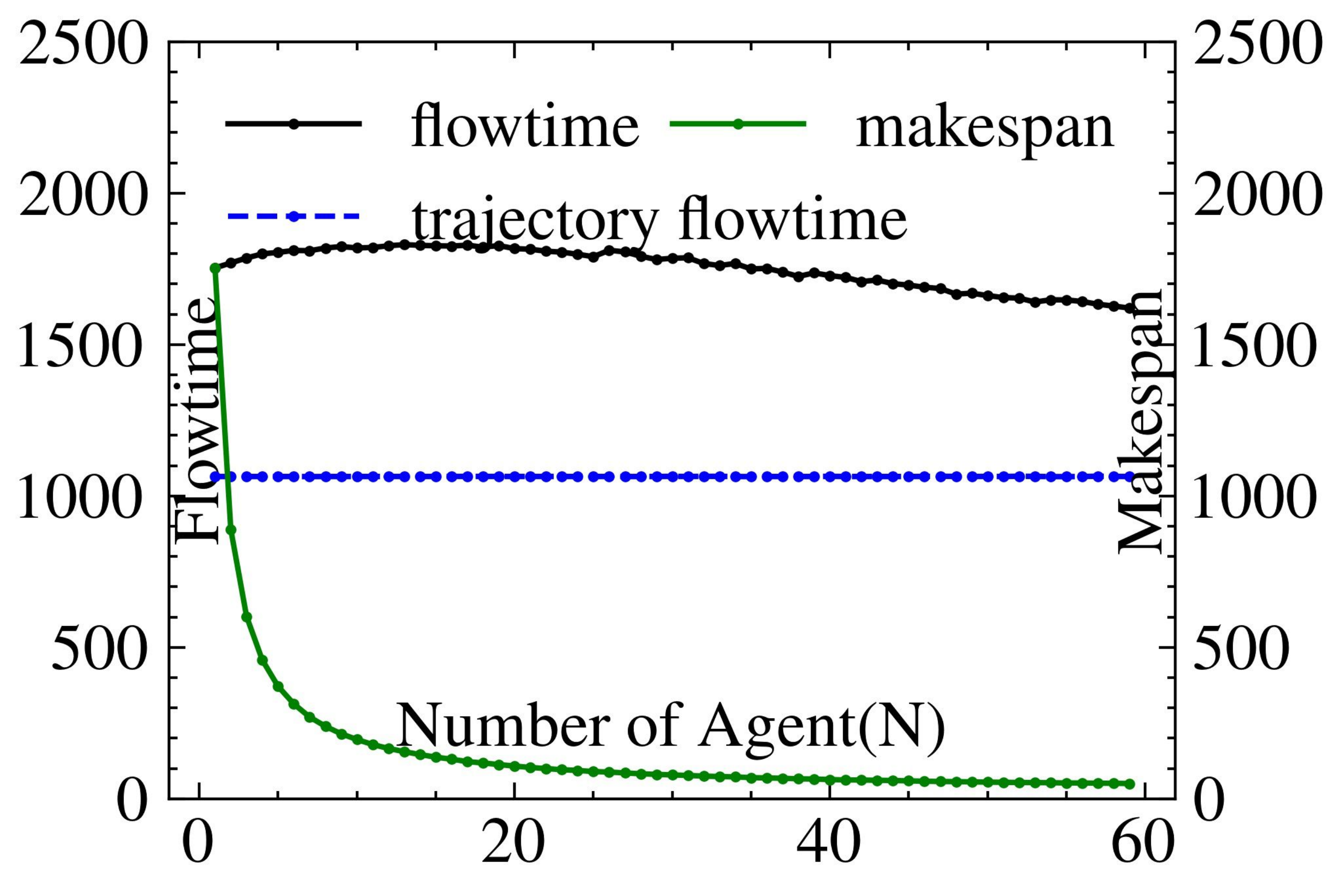}~~~~~~~~\includegraphics[height=85pt]{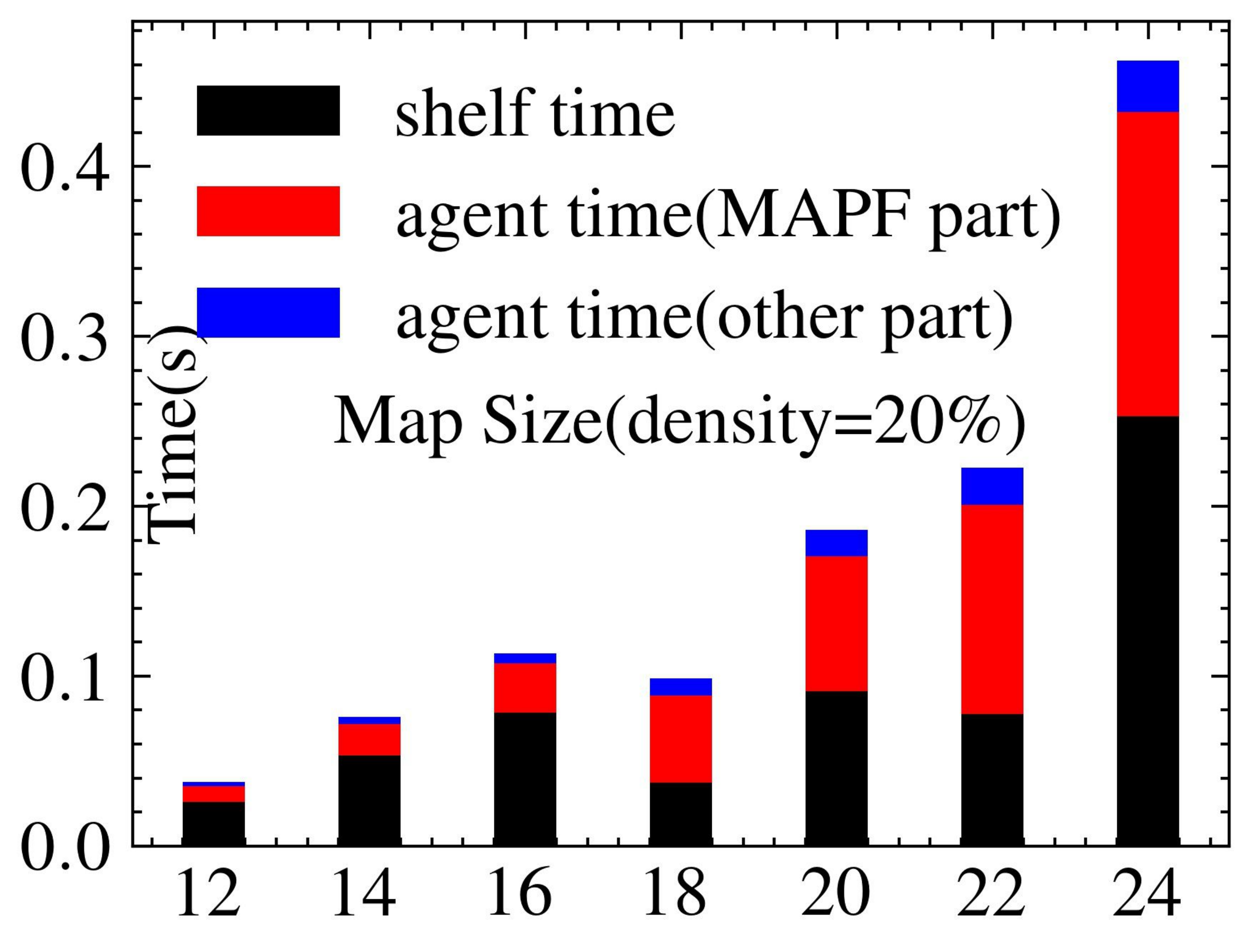}
    \caption{Results for IVF (den=20\%). Left: Flowtime ratios against trajectory flowtimes for 4 agents and different sizes. Middle: Effectiveness for different numbers of agents and size 24. Right: Time breakdown for different sizes and 8 agents.}
    \label{fig_dif_perf}
\end{figure*} 

\section{Experiments}

We conduct our experiments on a 3.1GHz Intel Core i5 laptop with 16GB RAM. We implement three variants of MAPF-DECOMP: MAPF-DECOMP without the IVF optimization, MAPF-DECOMP with IVF, and MAPF-DECOMP(PP) (labeled \textbf{\texttt{NIVF}}, \textbf{\texttt{IVF}}, and \textbf{\texttt{PP}}, respectively). They all use EECBS \cite{LiAAAI21a} to compute both trajectories for shelves and paths for agents and are implemented in C++. We adapt EECBS to compute safe 1-robust shelf trajectories for PP. We label IVF executions that use 1-robust shelf trajectories returned by EECBS as \textbf{\texttt{IVF-R}}.
The suboptimality factor $\omega$ of EECBS ranges from 1.2 to 1.8 for different settings to balance the effectiveness and efficiency.
We also implement two baseline algorithms: one that uses a single agent to execute the 1-robust shelf trajectories returned by EECBS in locked timesteps (labeled \textbf{\texttt{BASE}}), and another that used a single agent to execute 1-robust shelf trajectories returned by Push and Swap \cite{PushAndSwap} (labeled \textbf{\texttt{PAS}}). Unlike the trajectories returned by EECBS, the sequential 1-robust shelf trajectories returned by Push and Swap often have segments longer than one step, and PAS interprets the trajectories as a total order on the segments.

\subsection{MAPF-DECOMP with IVF}

We construct DD-MAPD instances on $n\times n$ square 2D 4-neighbor grids of different sizes (labeled \textbf{\texttt{size $n$}}) by randomly sampling blocks of $2\times 2$ cells as pickup locations of shelves until the pickup locations reach a certain density (percentage of all cells, labeled  \textbf{\texttt{den}}). We sample $0.1\cdot n^2$ shelves from all the shelves as the only ones that need relocation and sample their delivery locations from non-pickup cells. We randomly sample the start locations of the agents.
We evaluate 50 random instances per setting and report the mean over all solved instances.

\begin{table}[t]
    \renewcommand{\arraystretch}{0.5} 
    \centering
    \Huge\scalebox{0.3}{%
    \begin{tabular}{c|c|c|c|r|r|r|r|c|c}
    \toprule
    size & den & $M$ & $N$ & makespan & \makecell{flowtime} & \makecell{total\\time\\(s)} & \makecell{agent\\time\\(ms)} & $\omega$& succ\\
    \midrule\midrule
    8 & \multirow{6}{*}{40\%} & 25 & 4 &  31.10 & 109.32 & 1.81 & 2.82 & 1.2 & \multirow{6}{*}{\rotatebox{-90}{100\%}}\\
    \cline{1-1}\cline{3-9}
    \multirow{2}{*}{10} &      & \multirow{2}{*}{40} & 4 &  55.02 & 204.10 & 3.41 & 5.83 & \multirow{2}{*}{1.4} &\\
     &      &      & 8 &  31.83 & 209.26 & 3.39 & 8.97 &  &\\
    \cline{1-1}\cline{3-9}
    \multirow{2}{*}{12} &      & \multirow{2}{*}{57} & 4 &  89.18 & 339.53 & 0.30 & 11.81 & \multirow{2}{*}{1.6} &\\
     &      &      & 8 &  50.27 & 347.39 & 0.31 & 17.52 &  &\\
    \cline{1-1}\cline{3-9}
    16 &      & 102 & 8 & 105.68 & 772.60 & 0.92 & 55.83 & 1.8  &\\
    \midrule\midrule
    \multirow{2}{*}{16} & \multirow{7}{*}{20\%} & \multirow{2}{*}{51} & 4 & 140.98 & 534.94 & 0.10 & 22.51 & \multirow{2}{*}{1.2} & \multirow{7}{*}{\rotatebox{-90}{100\%}}\\
     &      &      & 8 & 74.34 & 530.60 & 0.11 & 34.82 &  &\\
    \cline{1-1}\cline{3-9}
    \multirow{2}{*}{24} &      & \multirow{2}{*}{115} & 4 & 452.69 & 1,775.31 & 0.38 & 134.82 & \multirow{2}{*}{1.2} &\\
     &      &      & 8 & 236.65 & 1,786.88 & 0.46 & 209.49 &  &\\
    \cline{1-1}\cline{3-9}
    \multirow{2}{*}{32} &      & \multirow{2}{*}{204} & 4 & 1,072.18 & 4,247.16 & 0.67 & 487.65 & \multirow{2}{*}{1.4} &\\
     &      &      & 8 & 546.39 & 4,249.12 & 0.92 & 737.61 &  &\\
    \cline{1-1}\cline{3-9}
    40 &      & 320 & 8 & 1,072.35 & 8,444.20 & 2.62 & 2,119.00 & 1.6  &\\
    \midrule\midrule\midrule
    \multicolumn{7}{c|}{\textbf{large-size instances, agent time reported in seconds} } & \makecell{(s)} & & \\
    \midrule\midrule
    \multirow{2}{*}{48} &   \multirow{8}{*}{20\%}   & \multirow{2}{*}{460} & 8 & 1,850.22 & 14,629.29 & 6.24 & 5.04 & \multirow{2}{*}{1.6}  & 98\%\\
     &      &  & 32 & 487.78 & 14,578.45 & 13.09 & 11.83 &   & 98\%\\
    \cline{1-1}\cline{3-10}
    \multirow{4}{*}{64} &      & \multirow{4}{*}{819} & 8 & 4,451.22 & 35,425.22 & 32.02 & 25.68 & \multirow{4}{*}{1.8}  & 98\%\\
    &  &  & 32 & 1,148.41 & 35,476.73 & 106.00 & 99.26 &  & 98\% \\
     &&  & 100 & 397.68 & 33,648.55 & 88.95 & 82.57 &  & 94\% \\
     &&  & 400 & 153.31 & 26,969.08 & 182.46 & 176.72 &  & 96\% \\
    \cline{1-1}\cline{3-10}
    \multirow{2}{*}{96}& & \multirow{2}{*}{1,843} & 32 & 3,909.00 & 123,383.24 & 302.92 & 270.27 & \multirow{2}{*}{1.8} & 92\% \\
     &&  & 100 & 1,264.05 & 118,965.28 & 1,060.16 & 1,019.93 &  & 80\% \\
     \midrule\midrule
    \bottomrule    
    \end{tabular}
    }
    \scalebox{0.3}{%
    \begin{tabular}{c|c|c|c|c|c|c|c}
    \toprule
    \multicolumn{8}{c}{\textbf{reasons for failed large-size instances}}\\
    \midrule
    size & den & $M$ & $N$ & instances & \makecell{(a) timeout} & \makecell{(b) small $N$} & \makecell{(c) incomplete} \\
    \midrule\midrule
    \multirow{2}{*}{48} &   \multirow{8}{*}{20\%}   & \multirow{2}{*}{460} & 8 & 1/50&0&1&0\\
     &      &  & 32 & 1/50&0&1&0\\
    \cline{1-1}\cline{3-8}
    \multirow{4}{*}{64} &      & \multirow{4}{*}{819} & 8 & 1/50&0&1&0\\
    &  &  & 32 &  1/50&0&1&0\\
     &&  & 100 &  3/50&0&3&0\\
     &&  & 400 &  2/50&0&0&2\\
    \cline{1-1}\cline{3-8}
    \multirow{2}{*}{96}& & \multirow{2}{*}{1,843} & 32 &  4/50&1&3&0\\
     &&  & 100 & 10/50&8&2&0\\
    \midrule
    \multicolumn{4}{c|}{sum}&23/400&9&8&2\\
    \bottomrule    
    \end{tabular}
    }
    \caption{Results for IVF in different settings.}
    \label{table_res}
\end{table}

\noindent\textbf{Optimization and numbers of future timesteps.} \Cref{fig_inv_step} shows the results for IVF when different values of $K$ (numbers of future timesteps) are used, where $K=\textnormal{inf}$ means that all active agents are included in the shelf assignment. The makespan tends to be smaller for larger $K$. $K=\textnormal{inf}$ does not necessarily result in the smallest makespan, where the last call of EECBS returns paths for a few agents that are much longer than for other paths since EECBS optimizes only the flowtime. As $K$ grows, the runtime excluding shelf trajectory computation (labeled \textbf{\texttt{agent time}}) (1) first drops since the makespan also drops and fewer calls to EECBS are made and (2) can then go up since each call to EECBS involves more agents. We use $K=8$ for IVF in all the following experiments to balance the runtime and the effectiveness. IVF (with $K>0$) is always more effective than NIVF due to better shelf assignments as a result of involving more agents. 

\noindent\textbf{Grid sizes, agent numbers, and shelf densities.}
Table \ref{table_res} (Top) shows that IVF solves all instances in seconds for small numbers of agents and that doubling the number of agents sometimes increases the flowtime, which sums up the completion times of all agents, but always reduces the makespan significantly. The agent times are large for large makespans due to the large numbers of calls of EECBS by AssignAndPlan(). Table \ref{table_res} (Middle) shows that IVF achieves high success rates (labeled \textbf{\texttt{succ}}) for hundreds of agents and more than one thousand shelves.
The total runtimes remain in a few minutes even though the task-assignment and path-planning functions must be executed many times for makespans of thousands of timesteps. 
Table \ref{table_res} (Bottom) categorizes the reasons for failed instances: (a) A timeout (1 min) for EECBS to compute shelf trajectories (typically for very large $M$); (b) Not enough agents to simultaneously move all shelves in a soft dependency cycle (typically for large $M$ and small $N$), which can be addressed by computing a 1-robust MAPF solution for shelves or using more agents by setting $N$ to be the number of shelves in the largest cycle; (c) The incompleteness of path planning for free agents in AssignAndPlan() (typically for very large $N$), which can be addressed by using deadlock-avoidance techniques if the given DD-MAPD instance is well-formed.

\noindent\textbf{Effectiveness.}
Figure \ref{fig_dif_perf} (Left) shows that the flowtime of the shelf trajectories (the sum of all trajectory lengths), which is a (trivial) lower bound on the flowtime and for which MAPF-DECOMP does not optimize, consistently contributes to a large portion ($\ge$60\%) of the flowtime across different instance sizes, which indicates that the decomposition and the execution of the trajectories of our framework are both effective.
Figure \ref{fig_dif_perf} (Middle) shows that to execute the same trajectory, as the number of agents increases, the flow time drops for large numbers of agents and the makespan drops in all cases. This is so because shelves are often assigned to close-by agents when there are many agents.

\noindent\textbf{Runtime breakdown.}
\Cref{fig_dif_perf} (Right) confirms that the runtime used to compute the trajectories of shelves (labeled \textbf{\texttt{shelf time}}) contributes to a large portion of the total runtime. It also suggests that an improvement in the efficiency of the MAPF solver would directly result in an improvement in the efficiency of our framework since most of the runtime is used by the MAPF solver.

\begin{table}[t]
    \renewcommand{\arraystretch}{0.5} 
    \centering
    \Huge\scalebox{0.3}{%
    \begin{tabular}{c|c|c|c|r|r|r|r|c|c}
    \toprule
    size  & $M$ & $N$ & algo. & makespan & \makecell{flowtime} & \makecell{total\\time\\(s)} & \makecell{agent\\time\\(s)} & $\omega$& succ\\
    \midrule\midrule
    \multirow{9}{*}{48} & \multirow{9}{*}{460}  & \multirow{3}{*}{1} &BASE& 190,243.52 &-& 1.20 & - &\multirow{9}{*}{1.6}&100\%\\
    &&&PAS& 14,707.44 &-& 2.31 & - &&100\%\\
    &&&IVF-R& 13,439.26 &-& 2.55 & 1.35 &&100\%\\
    \cline{3-8}\cline{10-10}
    && \multirow{3}{*}{8} & PP&   1,757.33& 13,910.77 & 2.11 & 0.93 && 100\%\\
    &&&  IVF-R&    1,750.96& 13,862.56 & 6.35 & 5.20 && 100\%\\
    &&&   IVF&   1,745.44& 13,812.10 & 6.86 & 5.80 && 96\%\\
    \cline{3-8}\cline{10-10}
    && \multirow{3}{*}{32}& PP&      463.71& 13,836.77 & 2.94& 1.58 && 100\%\\
    &&& IVF-R&      460.75& 13,705.56 & 10.68& 9.48 && 100\%\\
    &&& IVF&      461.35& 13,740.19 & 10.05& 9.13 && 96\%\\
    \midrule\midrule
    \multirow{12}{*}{64} & \multirow{12}{*}{819}  & \multirow{3}{*}{1} &BASE& 605,647.28 &-& 4.34 & - &\multirow{12}{*}{1.8}&100\%\\
    &&&PAS& 34,642.68 &-& 9.17 & - &&100\%\\
    &&&IVF-R& 32,560.80&-& 9.06 & 4.72 &&100\%\\
    \cline{3-8}\cline{10-10}
    && \multirow{3}{*}{8} & PP&   4,258.04& 33,877.76&  7.66 &  3.18 && 100\%\\
    &&& IVF-R&   4,254.92&33,839.10& 25.64 & 21.01 &  & 100\%\\
    &&& IVF&   4,256.45& 33,865.43& 24.43 & 20.98 && 98\%\\
    \cline{3-8}\cline{10-10}
    && \multirow{3}{*}{32}& PP& 1,105.00 & 34,090.16 & 10.69 & 5.38 && 100\%\\
    &&&  IVF-R& 1,099.69 & 33,943.73 & 41.29 & 36.64 && 100\%\\
    &&&   IVF& 1,096.71 & 33,835.06 & 40.50 & 37.01 && 98\%\\
    \cline{3-8}\cline{10-10}
    && \multirow{3}{*}{100} &PP& 384.81 & 32,695.28 & 17.78 & 11.00 && 100\%\\
    &&&IVF-R& 381.58 & 32,419.40 & 63.19 & 58.75 && 98\%\\
    &&&IVF&384.00 & 32,530.45 & 66.11 & 62.59 && 98\%\\
    \midrule\midrule
    \multirow{9}{*}{96}&\multirow{9}{*}{1,843} & \multirow{3}{*}{1} &BASE&  3,027,935.92 &-& 37.13 & - &\multirow{9}{*}{1.8}&100\%\\
    &&&PAS& 112,282.98 &-& 71.02 & - &&100\%\\
    &&&IVF-R& 111,258.48&-& 72.10 & 34.97 &&100\%\\
    \cline{3-8}\cline{10-10}
    &&\multirow{3}{*}{32}& PP& 3,817.24 & 120,368.60 & 79.49 & 33.06 && 100\%\\
    &&& IVF-R& 3,817.67 & 120,410.38 & 301.82 & 264.46 & & 98\%\\
    &&& IVF& 3,807.28 & 120,134.72 & 294.06 & 271.03 && 92\%\\
    \cline{3-8}\cline{10-10}
    && \multirow{3}{*}{100} &PP& 1,235.37 & 115,719.63 & 123.86 & 64.49 &  & 100\%\\
    &&&IVF-R& 1,228.53 & 115,274.80 & 623.54 & 585.69 &  & 98\%\\
    &&&IVF& 1,229.35 & 115,037.24 & 708.07 & 684.56 &  & 92\%\\
    \bottomrule    
    \end{tabular}
    }
    \caption{Results on large-size well-formed instances.}
    \label{table_new_multi}
\end{table}

\subsection{Comparison of Algorithms on Well-Formed Instances}

We follow a similar procedure to construct random DD-MAPD instances,
except that we do not place shelves along the perimeter of the grids but sample start locations of agents from cells on the perimeter, excluding the corners, to guarantee well-formedness.
\Cref{table_new_multi} shows that IVF-R with one agent is more effective than the two baselines and that using more agents results in further improvement. For example, the makespan for 100 agents is only 1.1\% of that for a single agent for instances of size 96. IVF-R and IVF tend to be more effective but are less efficient than PP since PP assigns tasks and plans paths for agents one at a time. PP solves all well-formed instances as expected. IVF-R has higher success rates than IVF since it does not fail for Reason (b) observed in the previous experiments but still fails for three instances in total due to the incompleteness of AssignAndPlan(). All algorithms do not time out with a one-minute runtime limit for each call to the MAPF solver. Overall, PP appears to be the the optimal choice of algorithm in practice if well-formedness is guaranteed since it strikes a good balance between efficiency and effectiveness.

\subsection{Warehouse Rearrangement Demo}

We construct 50 well-formed instances on 2D 4-neighbor grids of size $27\times 27$, with 32 agents starting along the perimeter, based on the layout of a fulfillment center comprising $8\times 4$ blocks of $5\times 2$ shelves. The delivery locations of these 320 shelves are arranged in a diagonally symmetrical configuration and randomly shuffled. IVF-R successfully solves all instances with an average total time of 30.68s and an average agent time of 4.85s using EECBS with $\omega = 1.8$. A demo video showcasing the execution on one of the instances is available at: \url{https://youtu.be/WFPl3wKDXXY}.

\section{Conclusions and Future Work}

We proposed a new algorithmic framework for solving DD-MAPD to make MAPF and MAPD applicable to multi-robot shelf rearrangement problems in large-scale warehouses. In this paper, we focus on the efficiency of our framework without sacrificing much of its effectiveness.

We propose two future extensions to our framework: (1) We intend to improve its effectiveness by making its MAPF solving for shelf trajectories aware of the subsequent decomposition and planning. (2) We propose to plan paths for active agents instead of letting them follow shelf trajectories.

\bibliographystyle{IEEEtran}
\bibliography{aaai23}

\end{document}